\documentclass[sigconf]{acmart}

\usepackage{booktabs} % For formal tables
\usepackage{amsmath,amssymb,amsfonts}
\usepackage{algorithmic}
\usepackage{graphicx}
\usepackage{textcomp}
\usepackage{xcolor}

\usepackage[T1]{fontenc}

\usepackage{soul}
\usepackage[utf8]{inputenc}
\usepackage{caption}
\usepackage{multirow}
\usepackage{multicol}
\usepackage{graphicx}
\usepackage{bm}
\usepackage{threeparttable}
\usepackage{tabularx}
\usepackage{tabu}
\usepackage{amssymb}
\usepackage{adjustbox}
\usepackage{booktabs}
\usepackage{array}
\usepackage{subcaption}
\usepackage[title]{appendix}

\newcommand{\rom}[1]{\uppercase\expandafter{\romannumeral #1\relax}}
\newcommand{\roml}[1]{\lowercase\expandafter{\romannumeral #1\relax}}
\newcolumntype{Y}{>{\centering\arraybackslash}X}

\newcolumntype{L}[1]{>{\raggedright\let\newline\\\arraybackslash\hspace{0pt}}m{#1}}
\newcolumntype{C}[1]{>{\centering\let\newline\\\arraybackslash\hspace{0pt}}m{#1}}
\newcolumntype{R}[1]{>{\raggedleft\let\newline\\\arraybackslash\hspace{0pt}}m{#1}}

\def\BibTeX{{\rm B\kern-.05em{\sc i\kern-.025em b}\kern-.08em
    T\kern-.1667em\lower.7ex\hbox{E}\kern-.125emX}}

\setcopyright{rightsretained}
\acmConference[LAK'2019]{Learning Analytics and Knowledge Conference}{March 2019}{Tempe, Arizona}
\acmYear{2019}
\copyrightyear{2019}
\acmPrice{15.00}

\begin{document}

\title{Reliable Deep Grade Prediction with Uncertainty Estimation}

\author{Qian Hu}
% \authornote{Dr.~Trovato insisted his name be first.}
% \orcid{1234-5678-9012}
\affiliation{%
  \institution{George Mason University}
  \streetaddress{4400 University Drive}
  \city{Fairfax}
  \state{Virginia}
  \postcode{22030}
}
\email{qhu3@gmu.edu}

\author{Huzefa Rangwala}
% \authornote{The secretary disavows any knowledge of this author's actions.}
\affiliation{%
  \institution{George Mason University}
  \streetaddress{4400 University Drive}
  \city{Fairfax}
  \state{Vifginia}
  \postcode{22030}
}
\email{rangwala@cs.gmu.edu}

\begin{abstract}
Currently, college-going students are taking longer to graduate than their parental generations. Further, in the United States, the six-year graduation rate has been 59\% for decades. Improving the educational quality by training better-prepared students who can successfully graduate in a timely manner is critical. Accurately predicting students' grades in future courses has attracted much attention as it can help identify at-risk students early so that personalized
feedback can be provided to them on time by advisors. 
Prior research on students' grade prediction include shallow linear models; however, students' learning is a highly complex process that involves the accumulation of knowledge across a 
sequence of courses that can not be sufficiently modeled by these linear models.  In addition to that, prior approaches focus on prediction accuracy without considering prediction uncertainty, 
which is essential for advising and decision making. In this work, we present two types of Bayesian deep 
learning models for grade prediction under a course-specific framework: 
i)Multilayer Perceptron (MLP) and ii) Recurrent Neural Network (RNN). 
These course-specific models are based on the assumption that prior courses can provide students with knowledge for future courses so that grades of prior courses can be used to predict grades in a future course. 
The MLP ignores the temporal dynamics of students' knowledge evolution. 
Hence, we propose RNN for students' performance prediction. To evaluate the performance of the proposed models, we performed extensive experiments on data collected from a large public university. The experimental results show that the proposed models achieve better performance than prior state-of-the-art approaches. Besides more accurate results, Bayesian deep learning models estimate uncertainty associated with the predictions. We explore how uncertainty estimation
can be applied towards developing a reliable educational early 
warning system. In addition to uncertainty, we also develop an approach 
to explain the prediction results, which is useful for advisors to 
provide personalized feedback to students.
\end{abstract}

\begin{CCSXML}
<ccs2012>
<concept>
<concept_id>10010405.10010489.10010490</concept_id>
<concept_desc>Applied computing~Computer-assisted instruction</concept_desc>
<concept_significance>500</concept_significance>
</concept>
<concept>
<concept_id>10010405.10010489.10010495</concept_id>
<concept_desc>Applied computing~E-learning</concept_desc>
<concept_significance>300</concept_significance>
</concept>
<concept>
<concept_id>10002951.10003227.10003351</concept_id>
<concept_desc>Information systems~Data mining</concept_desc>
<concept_significance>300</concept_significance>
</concept>
<concept>
<concept_id>10010147.10010257.10010258.10010259</concept_id>
<concept_desc>Computing methodologies~Supervised learning</concept_desc>
<concept_significance>300</concept_significance>
</concept>
</ccs2012>
\end{CCSXML}

\ccsdesc[500]{Applied computing~Computer-assisted instruction}
\ccsdesc[300]{Applied computing~E-learning}
\ccsdesc[300]{Information systems~Data mining}
\ccsdesc[300]{Computing methodologies~Supervised learning}

\keywords{Bayesian Deep Learning, Uncertainty, Sequential Models, Educational Data Mining, Grade Prediction}

\copyrightyear{2019} 
\acmYear{2019} 
\setcopyright{acmcopyright}
\acmConference[LAK19]{The 9th International Learning Analytics \& Knowledge Conference}{March 4--8, 2019}{Tempe, AZ, USA}
\acmBooktitle{The 9th International Learning Analytics \& Knowledge Conference (LAK19), March 4--8, 2019, Tempe, AZ, USA}
\acmPrice{15.00}
\acmDOI{10.1145/3303772.3303802}
\acmISBN{978-1-4503-6256-6/19/03}

\maketitle

\section{Introduction}
The average six-year graduation rate
for undergraduate programs in the United States 
has been around 
59\% for over a decade \cite{stillwell2013public}. More than half of 
the graduating 
students take six years to finish four-year programs. The additional time 
required by students and low graduation rates  has  high 
human, monetary and societal costs with regards to 
workforce training and economic growth.  Lack of  proper academic 
preparation  
and planning are some of the 
main reasons  that lead to student failure in higher education \cite{four_year_myth}. To improve  
retention rates 
and help 
students graduate in a timely manner, we aim to develop analytics-driven
early warning and degree planning systems 
that can identify at-risk students; so that advisors can provide
them timely and personalized feedback/advice. Grade prediction is 
fundamental for these systems. 

Prior next-term grade 
prediction methods usually train 
a one-size-fits-all model that 
predicts students' grades in multiple courses \cite{polyzou2016grade}. However, different 
courses have different characteristics such as prerequisites, knowledge content, instructors and difficulty level.  To address 
this problem, Polyzou et al. proposed a 
course-specific framework, which was shown to be successful for accurately predicting students' grades \cite{polyzou2016grade,polyzou2016grade1}. Course-specific methods 
identify a subsets of prior courses on a course-by-course basis to predict a student's grade in a target course. They are based on the
assumption that students 
accumulate necessary knowledge to take future courses by taking a sequence of prior courses. Our models are based upon this course-specific framework.

From the perspective of educational psychology, learning is affected by both external and internal 
factors such as motivation, study habits, attention and instructor pedagogy \cite{mangal2002advanced,benjamin2014factors}, which 
bring about challenges for grade prediction. These challenges are further exacerbated by the fact that learning is a 
reflection of human cognition which is a complex process \cite{piech2015deep}. Existing course-specific models are linear shallow learners, e.g. linear regression or low-rank matrix factorization. These shallow learners may 
not be capable of capturing complex interactions  underlying student's learning. To better model 
students' learning process,
% and improve prediction accuracy, 
we propose to use deep learning models.
Another drawback of traditional grade prediction methods is that the predicted grade is a point estimation. To make informed decision based on the predicted results, we need to 
know if the prediction system is confident or not. If the system is confident enough about the predictions, we can rely on them and take corresponding actions. However, if the prediction is not reliable, human advisors should decide what to do. 
Compared to traditional deep learning models, Bayesian deep learning models can provide principled uncertainty estimation.

% for modeling student's performance. 
Specifically, we propose two types of Bayesian deep learning models, (\rom{1}) Multilayer Perceptron (MLP) \cite{rosenblatt1961principles}, (\rom{2}) Long Short Term Memory (LSTM) networks \cite{hochreiter1997long}. MLP consists of
hierarchical hidden layers that maps the input vector to an output target. The input vector is treated as static and hence the temporal dynamics of the input 
data are ignored by MLP. 
% However, the students take courses in a sequential manner. 
To capture student's knowledge evolution, we also propose a LSTM model. Theoretically, RNNs are able to model arbitrarily long sequential data. However, in practice, because of the vanishing gradient problem, vanilla RNNs fail to capture long-term dependencies. For grade prediction, a course taken several semesters ago might still have influence on the student's performance in a future course.
% , such as the prerequisites. 
To model such long-term dependencies, we choose to use LSTM model.

The proposed models are evaluated on datasets 
extracted from University X by using different evaluation
metrics. The results show that the proposed models 
outperform the comparative state-of-the-art methods in all aspects. 
%
%When a model is used for decision making, we can not take the results blindly. Uncertainty and interpretability are essential for decision making.
To trust the predictions from a model, we need to know if the system is confident about its predictions 
or not. 
%Bayesian models
%provide confidence estimates with their prediction results.  
We provide empirical 
results about model uncertainty and investigate case studies towards developing a 
reliable educational early warning system. We also propose a method to explain the models' predictions, which 
identifies a list of influential prior courses that 
lead to a student's failure in the target course. 
%
%The influential prior courses can be used by the advisors 
%to provide personalized advising to students.

The main contributions of this work can be summarized as follows:
\begin{itemize}
    \item We propose two types of course-specific Bayesian deep learning models for grade prediction, namely, course-specific MLP and LSTM. Compared to existing methods, the proposed models have better modeling capability and prediction accuracy.
    \item The proposed models can provide prediction uncertainty which is essential for decision making. Based on uncertainty estimation, we show how uncertainty can help build a reliable educational early warning system.
    \item In addition to uncertainty estimation, we propose a method to explain the prediction results, which can identify influential courses that results in a student's failure of a course.
    \item We propose a method to evaluate the models' capability of catching at-risk students. The evaluation results show that the proposed methods outperform several baseline methods for this task.
\end{itemize}

\section{Related Work}
The application of analytics to improve educational 
quality can be seen in many areas related to modeling of learners 
\cite{lan2014sparse}, predicting and advising learners 
\cite{elbadrawy2016predicting}, automated content enhancement 
\cite{agrawal2014mining}, knowledge tracing 
\cite{yudelson2013individualized,corbett1994knowledge} and course/topic 
recommendations to students \cite{elbadrawy2016domain}. Among them, 
student's academic performance prediction has attracted much attention, 
as it underlies applications to several AI-based decision making 
systems including educational early warning systems, degree planning 
and academic trajectory planning \cite{elbadrawy2016predicting}.
In light of this paper's scope, we only review approaches for 
student's performance prediction and predictive uncertainty estimation.

\subsection{Student Performance Prediction}
Several machine learning algorithms have been applied to tackle the student 
performance prediction problem \cite{bydvzovska2015student,sorour2015student}. 
Al-Barak et al. applied decision trees for grade prediction by using 
students' transcript data \cite{al2016predicting}. Umair et al. used 
Support Vector Machines (SVMs) to select key training instances for 
grade prediction \cite{umair2018predicting}. Recommender systems 
based methods including collaborative filtering 
\cite{bydvzovska2015collaborative}, matrix factorization 
\cite{koren2009matrix} and factorization machines \cite{sweeney2016next} 
have been proposed for grade prediction. These approaches use a 
one-size-fits-all framework for training the model and prediction. 
Polyzou et al. proposed a personalized model that is specific to each 
course and student \cite{polyzou2016grade1}. Student-course enrollment 
patterns have grouping structures which result in missing not at 
random patterns of student grade data. Leveraging this, 
Elbadrawy et al. proposed a domain-aware grade prediction algorithm 
for student's performance prediction and course recommendation 
\cite{elbadrawy2016domain}. Since students accumulate knowledge 
by taking courses sequentially within the academic programs, 
it is assumed that the knowledge state of the students is evolving.  
Ren et al. proposed a temporal course-wise influence model which 
incorporates the influence of prior courses in a sequential way, 
however, up to two terms \cite{ren2017grade}.

% \vspace{-3mm} 
Several works use deep learning to model student learning habits 
and predict performance. Livieris et al. developed a neural network 
based classifier to predict whether a student will have poor 
performance in a Math course \cite{livieris2012predicting}. 
Gedeon et al. trained a feedforward neural network to predict a 
student's final grade in a computer science course using data from 
teaching sessions and provided interpretability of the prediction 
results by generating a set of rules \cite{gedeon1993explaining}. 
Yang et al. designed a time series neural network using a student's 
clickstream data while watching video lectures in massive open online 
courses (MOOCs) \cite{yang2017behavior}. Okubo et al. proposed a 
recurrent neural network classifier to predict a student's grade by 
using data from various logs of learning activities \cite{okubostudents}. 
For modeling student's learning process within Intelligent Tutoring Systems (ITS), 
Piech et al. proposed using deep knowledge tracing \cite{piech2015deep}. Most of the proposed 
neural network models  were developed for in-class prediction or for 
intelligent tutoring systems that model student learning in a single course. 
The DKT models are similar to our proposed LSTM model; however, 
the DKT models only incorporate one response each time step. Our 
proposed LSTM model is more flexible and can incorporate several 
(responses) grades of prior courses taken together in the same semester.

The existing deep learning models either ignore the temporal dynamics 
of student's grade data or are designed only for a single course by 
using data within that single course. In this paper, our proposed 
models aim to predict student's performance by using data across 
several teaching sessions and the LSTM model can take into account 
the sequential aspect of a student accumulating knowledge across 
multiple semesters.

\subsection{Predictive Uncertainty}
\begin{figure}[h]
    \centering
    \begin{subfigure}[b]{.45\columnwidth}
        \centering
        \includegraphics[width=\linewidth]{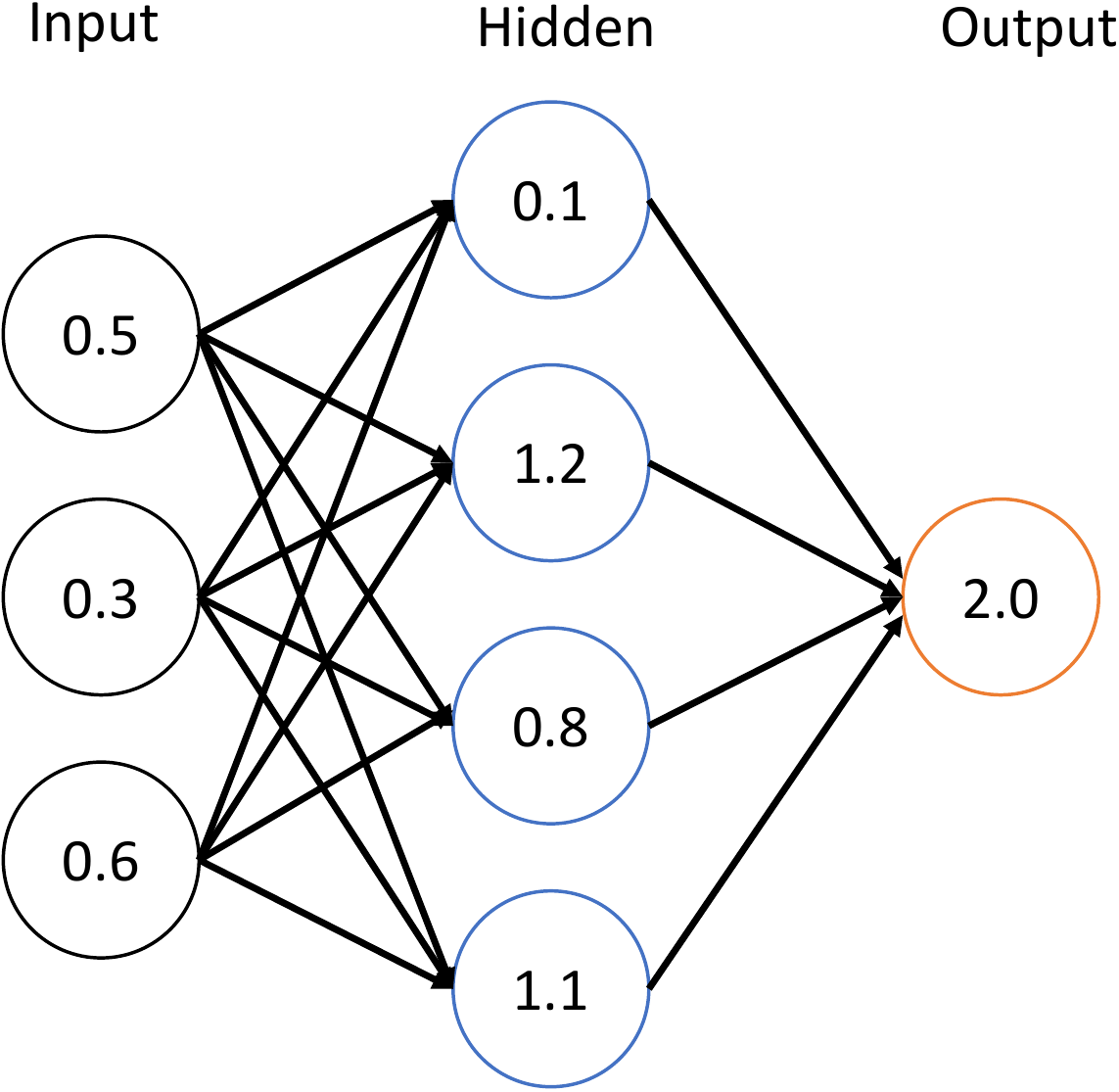}
        \caption{Neural Netowrk}
    \end{subfigure}
    \vfill
    \begin{subfigure}[b]{.45\columnwidth}
        \centering
        \includegraphics[width=\linewidth]{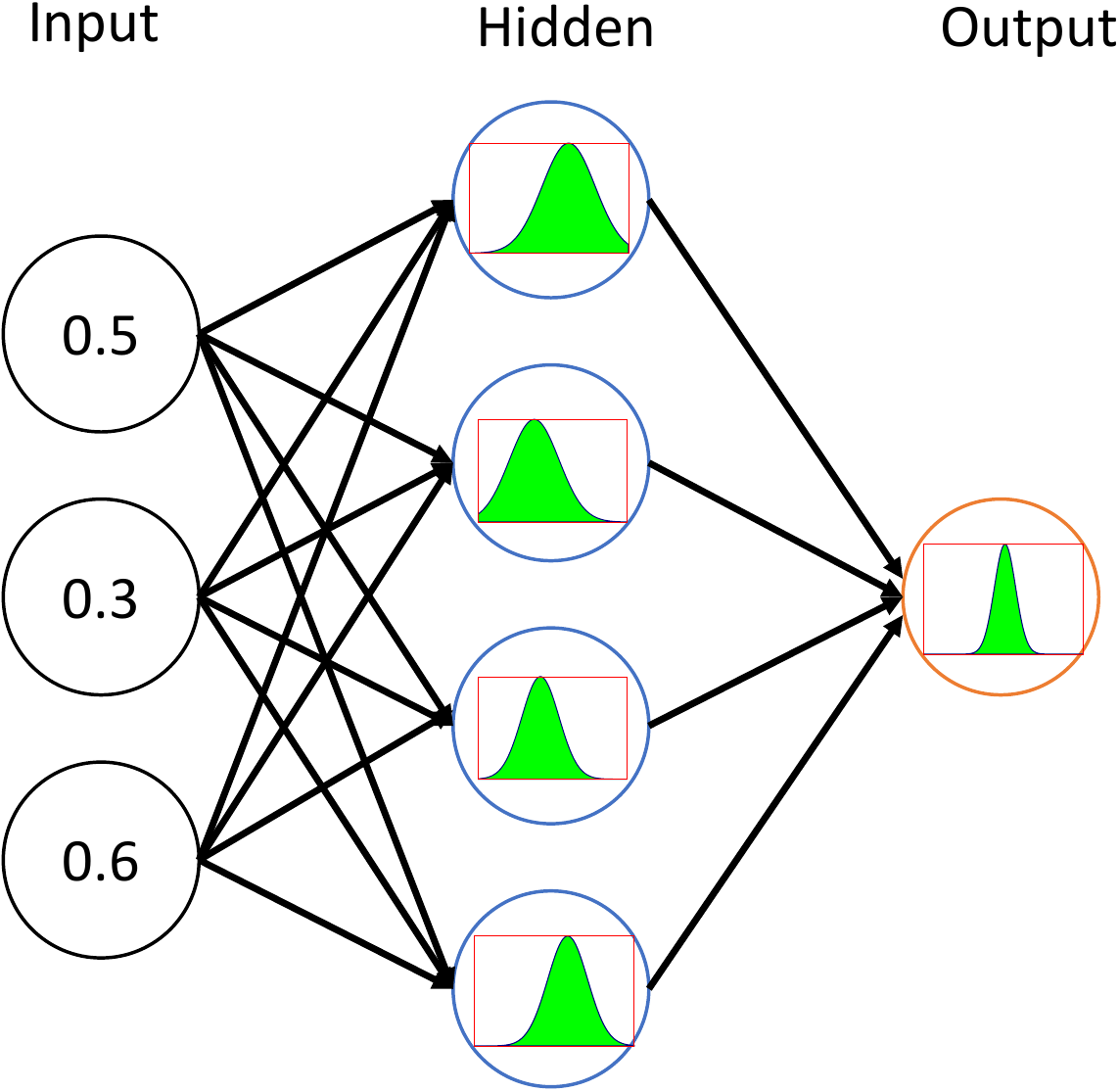}
        \caption{Bayesian Neural Network}
    \end{subfigure}
    \caption{Comparision of Neural Network and Bayesian Neural Network} \label{fig:bnn}
\end{figure}

Deep learning models have achieved state-of-the-art performance in 
many areas due to their abilities to model complex patterns \cite{lecun2015deep}. 
However, general deep learning models cannot represent uncertainty, 
which is critical for decision-making. Bayesian models have 
the advantage of providing principled uncertainty estimation. 
Therefore, combining Bayesian approaches with deep learning models 
is a way to obtain benefits from these two perspectives. Figure \ref{fig:bnn} 
shows the difference between a traditional 
deep learning (neural network) and Bayesian deep 
learning model. 

Bayesian deep learning models place a prior distribution over model 
parameters; the model is updated by Bayes' rule with observed data. 
The posterior distribution of the model parameters is the learned model. 
Due to possible non-linear activation functions that can be 
applied to neurons, exact 
model posterior is not available. Approximate inference methods are used for 
model training, such as variational inference \cite{gal2015modern}. However, these methods 
have a  high computational cost and are 
hard to scale 
in practice.

Recently, Monte Carlo (MC) dropout has
been proposed by Gal et al. \cite{gal2016dropout}, 
which is efficient for uncertainty estimation and requires no 
change in the designed 
model architecture. In this work, we adopt MC dropout 
as a way to estimate 
prediction uncertainty and explain the details in Section 
\ref{uncertainty_estimation}.

\section{Methods}

\subsection{Model Learning Framework}

Given records of $n$ students and $m$ courses, we extract 
the grades to form a sparse grade matrix $\mathbf{G} \in \mathbf{R}^{n \times m}$. 
In addition, we have the information associated with the semester (time) when the particular grade was 
obtained.  
Further, the data includes student-related features 
(e.g., academic level, previous GPAs, major, etc.) and course-related features 
(e.g., course level, discipline, credit hours, etc.). These content features 
are combined to form a feature vector associated with a student-course pair.

Given a student's grades in 
the  courses taken before the target course (referred to as prior courses), the 
objective of the next-term grade prediction problem is to predict the grade that the 
student will achieve in a course to be  taken in the next semester (term).
%
%Traditional grade prediction framework focus on predicting student's next-term GPA \cite{sweeney2016next} i.e., average grade points of courses to be taken in next term, by using grades from previous terms. The issue with this framework is that predicted grade for a specific course is not available. 
%
To predict grade in a course-wise manner, we adopt course-specific
framework \cite{polyzou2016grade1}. Under this framework, different models 
are learnt for different courses. To predict a student's grades in next courses, his/her grades from prior courses are fed into corresponding models.

% The training data for a MLP model $F^c$ of course $c$ is extracted as 
% following. First, we extract from the grade matrix $\mathbf{G}$ the students that have taken course $c$. For each of these students, their 
% grades corresponding to the courses taken prior to course $c$ are extracted to form a course-specific grade matrix $\mathbf{G}^c \in \mathbb{R}^{n_c \times m}$, where $n_c$ is the number of students 
% and $m$ is the number of prior courses. The grades of the $n_c$ students in course $c$ is 
% represented as $\mathbf{y}^c \in \mathbb{R}^{n_c}$. This course-specific training data extraction technique was first introduced by Polyzou et al. \cite{polyzou2016grade} and 
% is used for training the deep learning models proposed here. 

% Moved it back -- better sequencing. 

\subsection{Multilayer Perceptron}
Traditional grade prediction models are linear models, such as linear regression.
Compared to linear models, the key advantage
of multilayer perceptron
comes from its hierarchical hidden layers that capture complex
interactions and non-linearities. The theoretical foundation is given by the Kolmogorov-Arnold representation theorem \cite{arnold1957functions,kolmogorov1963representation}; every multivariate 
continuous function can be represented as a superposition of one-dimensional
continuous functions. 

Given an input vector $\bm{x}$, the task of the multilayer perceptron algorithm is to map $\bm{x}$ to output $\bm{y}$, which has the following form
\begin{equation}
\bm{y} = F(\bm{x})
\end{equation}

To estimate a student's grade in course $c$ by using the course-specifc MLP model $F^c$, we have
\begin{equation}
\hat{y}^c = F^c(\mathbf{s})
\end{equation}
where $\mathbf{s} \in \mathbb{R}^m$ is the vector of the student's grades in the prior courses.

\subsection{Long Short Term Memory}
\begin{figure}[h!]
	\centering
	\includegraphics[width=0.90\linewidth]{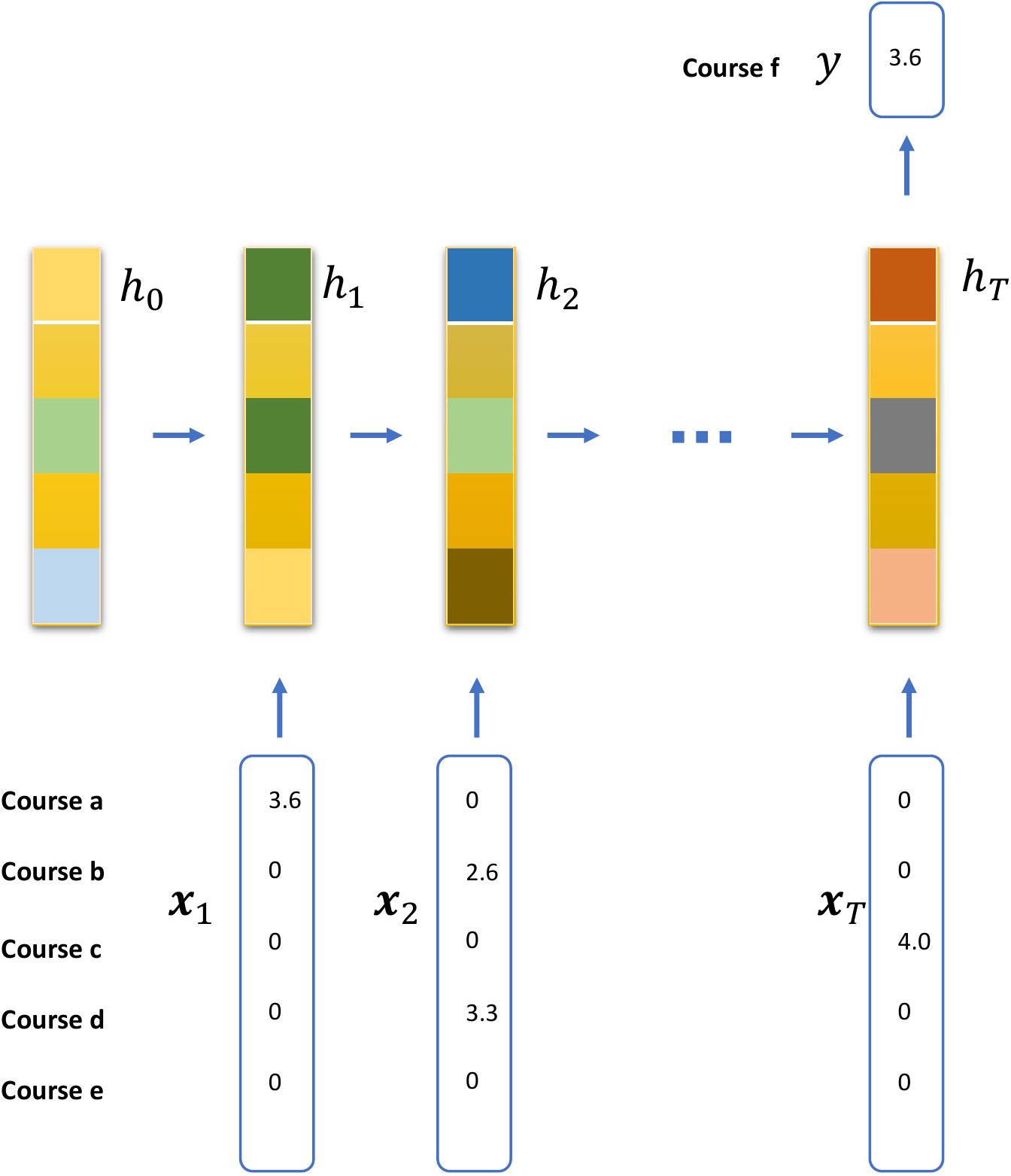}
	\caption{In this example, we want to predict a student's grade in target course $f$ by using grades of the courses taken prior to course $f$ include $a, b, c, d, e$. $\bf{x}_t$ represents the grades of courses in term $t$. $y$ represents the predicted grade. The student took courses ($a$), ($b$, $d$), ..., ($c$) in semester 1, 2, ..., $T$ and obtained (3.6), (2.6, 3.3), ..., (4.0) in this example,  respectively.} \label{fig:lstm}
\end{figure}

To capture the sequential characteristics of students' grades in prior courses, we model 
the learning behavior and performance using  recurrent neural networks with 
long short term memory \cite{hochreiter1997long} (LSTM). The standard RNN model has the vanishing gradient problem and is 
unable to capture long-range dependencies. In our case, an 
course taken several semesters before, such as a prerequisite, plays an 
important role in determining a student's performance in a target course. 
To solve the long-term dependency problem, LSTM is proposed for sequential data. 
The hidden states of LSTM capture the student's knowledge states, which models a 
student's knowledge evolution. The hidden states are updated as the student 
enrolls for courses and obtain grades in them. Figure \ref{fig:lstm} shows the LSTM approach for 
modeling student's learning process. At the beginning, a student has some prior knowledge before taking any courses; the student's 
knowledge states evolve as he/she take 
courses, as indicated by different colors at each time step in Figure \ref{fig:lstm}. A student's knowledge states 
influences
his/her performance in a course .The last
hidden state $\mathbf{h}_T$ is used to predict his/her grade in a target course within the course-specific framework.

LSTM is a gated recurrent neural network, which consists of forget gate and 
input gate. The forget gate decides which part of the 
information to forget from the cell state. This is useful when the 
same knowledge can be obtained by taking two different courses. A student's knowledge state corresponding to
knowledge acquired by taking the first course can be discarded while renewed by using the second one. When student 
takes 
a new course, his/her knowledge state is updated. In LSTM, this is done by the 
information layer and input gate; input gate decides which new information should be added into the cell state. The output from LSTM is hidden state which represent student's current knowledge state.

To estimate a student's grade in the target course by using LSTM model, we first extract the student's grades in the prior courses with timestamp i.e., in which terms the prior courses 
are taken. The grades in 
term $t$ are represented as multiple-hot encoded vector $\mathbf{x}_t$ --- as more than one course can be taken together in one semester --- where the entries of $\mathbf{x}_t$ corresponds to the grades of courses taken in semester $t$; $0$ represents the corresponding courses are not taken.  If the  grade obtained is 0 (F), we use a small number (0.1) to represent it to differentiate it from courses that are not taken. We input the sequence of the encoded vectors $\mathbf{x}_1, \mathbf{x}_2, ..., \mathbf{x}_T$ to the model and the hidden state from the last step $\mathbf{h}_T$ is fed into a fully connected layer, the output of which is the predicted grade:
\begin{equation}
	y = \mathbf{w}_{yh} \cdot \mathbf{h}_T + b_y
\end{equation}
where $\mathbf{h}_T$ is the last hidden state, $\mathbf{w}_{yh}$ is the parameters of the fully connected layer and $b_y$ is the bias term.

\begin{table*}[ht!]
    \centering
    \caption{Dataset Statistics} \label{tab:data}
    \begin{threeparttable}
    \begin{tabularx}{0.6\textwidth}{@{}Y|YYY|YYY@{}}
    \hline \hline
    \multirow{2}{2em}{Major}
    & \multicolumn{3}{c|}{Fall 2016} & \multicolumn{3}{c}{Spring 2017} \\
    \cline{2-7}
    & \#S & \#C & \#G & \#S & \#C & \#G \\
    \hline
    CS & 2,664 & 18 & 22,246 & 3,728 & 19 & 33,039 \\
    ECE & 1,160 & 16 & 16,415 & 1,421 & 15 & 23,459 \\
    BIOL & 2,736 & 19 & 20,984 & 6,002 & 20 & 42,895 \\
    PSYC & 2,980 & 20 & 14,966 & 4,628 & 20 & 23,560 \\
    CEIE & 1,525 & 18 & 23,954 & 1,873 & 17 & 28,198 \\
    \hline
    Overall & 11065 & 91 & 98,565 & 17652 & 91 & 151,151 \\
    \hline
    \end{tabularx}
    \begin{tablenotes}
        \scriptsize
        \item \#S number of students, \#C number of courses, \#G number of grades
        % \item 
        % \item \#G number of grades
    \end{tablenotes}
    \end{threeparttable}
\end{table*}

\subsection{Uncertainty Estimation} \label{uncertainty_estimation}
Given input data $\mathbf{x}$, the output of an Bayesian deep learning model $f(\cdot)$ is mean $\hat{y}$ and standard deviation $\sigma$, where $\sigma$ is treated as uncertainty; the lower the standard deviation, the higher the predictive confidence.
Bayesian models such as Gaussian Process provide principled uncertainty estimation, however, they are computationally prohibitive and hard to scale to large-scale datasets. Yarin et al., showed that dropout can be interpreted as a Bayesian approximation and Monte Carlo (MC) dropout is proposed to obtain prediction uncertainty \cite{gal2016dropout}. Dropout is first proposed as a method for preventing overfitting in neural networks. The basic idea of MC dropout is that for each input, we repeat the prediction for $T$ iterations to get $T$ different outputs, at each iteration neurons are randomly set to zero with some dropout probability. In the next section, we describe how to obtain uncertainty by using Monte Carlo dropout.

Given an deep learning model trained with dropout probability $p$, we sample T sets of model parameters ${W_1, W_2, ..., W_T}$ with different dropout masks to have different model realizations $f^{W_1},f^{W_2},...,f^{W_T}$. 
For an input $\mathbf{x}_i$, the outputs from $T$ model realizations are
\begin{equation}
    \hat{y}_i^t = f^{W_t}(\mathbf{x}_i)
\end{equation}
The prediction mean $\overline{y}$ is estimated as
\begin{equation}
    \overline{y} \approx \frac{1}{T}\sum_{t=1}^{T}\hat{y}_i^t
\end{equation}
The prediction variance is estimated as
\begin{equation}
    \sigma_i^2 = \tau^{-1} + \frac{1}{T}\sum_{t=1}^{T}(\hat{y}_i^t)^2 - \overline{y}^2
\end{equation}
which equals sample variance plus model uncertainty $\tau^{-1}$, where $\tau$ is a hyperparameter which needs to be tuned for different datasets \cite{gal2016dropout}.

Given prediction mean $\overline{y}$ and variance $\sigma_i^2$, an $\alpha$-level prediction interval is calculated as
\begin{equation}
    [\overline{y} - z_{a/2}\sigma, \; \overline{y} + z_{a/2}\sigma]
\end{equation}
where $z_{a/2}$ is the upper $(1-C)/2$ critical value for standard normal distribution. For example, $95\%$ prediction interval can be calculated as $[\overline{y} - 1.96\sigma, \; \overline{y} + 1.96\sigma]$.

\subsection{Interpretability} \label{explain}
When a model is used for decision making, it is necessary for practitioners to have confidence in the predictions in order to act upon them~\cite{ribeiro2016should,li2016understanding,avati2017improving}. When an instructor is notified of an
at-risk student, they need to know not only the predicted grades but also the reasons associated with 
the corresponding predictions (e.g., which prior courses lead to a student's failure in the target course). Towards this end, we develop an approach to explain the predictions made by the proposed model. The course-specific model 
assumes that the 
knowledge needed for a course is accumulated when taking the prior courses. As such, one of the factors associated with 
student's performance is his grade/performance in the prior courses. We compute the influence of a prior course 
in the following way. Given a trained model $M$ and a student $s$, the grade predicted by the model for this student is denoted as $\hat{y}_s$. Let $p$ be a prior course and $\hat{y}(\neg p)_s$ be the predicted grade if the corresponding grade of course $p$ is set to full grade, namely, 4.0 in the input to the model. For student $s$, the influence of course $c$ --- denoted by $I_c(s, p)$ --- is computed as
\begin{equation}
I_c(s, p) = \hat{y}(\neg p)_s - \hat{y}_s
\end{equation}

The intuition behind this approach is that if a student could have obtained higher grades 
in a prior course, he/she is likely to have better performance in the target course.
Based on this information a 
student could be advised to prepare or review the material in these 
influential courses so as to  be successful in the target course.    

This can also be used to improve the curriculum structure. 
By considering students collectively, if there exists 
a prior course that consistently has a high influence 
for a target course across several students, then this prior 
course material needs to be a prerequisite or reviewed in class (if not already present).
To compute the influence of a prior course on a target 
course, we observe that different students have a different grade in a prior course.  Instead of setting the grade of a 
prior course to 4.0, we increase its grade by a fixed value of 1.0 The following equations describe how to compute the influence of a prior course on a target course.

\begin{align}
    & I^*_c(s, p) = y^*(p_{+1.0}) - \hat{y}_s \\
    & I_c(p) = \sum_{s\in S} I^*_c(s, p)
\end{align}
where $p$ is the prior course, $S$ is all the students that have taken target course $c$, $y^*(p_{+1.0})$ is the predicted grade if grade of prior course $p$ is increased by 1.

\section{Experimental Protocols}
\subsection{Dataset Description} \label{data_description}
The methods are evaluated on a 
dataset from 
University X.
We choose the largest five 
undergraduate majors including: (\roml{1}) Computer Science (CS), (\roml{2}) Electrical Engineering (ECE), (\roml{3}) Biology (BIOL), (\roml{4}) Psychology (PSYC) and 
(\roml{5}) Civil Engineering (CEIE). To build a course-specific model for a target course, we choose the prior courses according to the University Catalog from Fall 2009 to Spring 2017.

% The models are evaluated so as to 
The evaluation
simulates the real-world scenario of
predicting the next-term grades for students.  Specifically, the models 
are trained on the data up to term $T - 1$ and tested on term $T$. The last two terms are chosen
as testing terms, i.e., Fall 2016 and Spring 2017. As an example, 
to evaluate the performance of predictions on term Fall 2016, the model is trained on data from Fall 2009 to Fall 2015; and data from Spring 2016 is 
used for selecting hyperparameters. Dataset statistics are in Table \ref{tab:data}.

\subsection{Model Training}
The deep learning models  are trained 
by using the 
Adam \cite{kingma2014adam} optimizer. For the hyperparameters, we use the 
grid search to choose the best combination on  the 
validation dataset as described above. Every 50 iterations, we take a snapshot of the model and the model that 
performs the 
best on the validation dataset is selected for final evaluation on the test set. 
The hyper-parameters for MLP include the 
number of layers (ranging from 2 to 10) and the number of neurons 
in each of the 
hidden layers (ranging from 2 - 50). For the stacked-LSTM, the parameters include the 
number of hidden dimensions (ranging from 10 to 100) and  the number of 
stacked layers (ranging from 1 to 5). The activation function is 
Rectified Linear Unit (ReLU) and the learning rate was set to 
0.001. The configuration parameters for Adam are set to 
default values  ($\beta_1$ is 0.9, $\beta_2$ is 0.999 and $\epsilon$ is 10e-8). 

\subsection{Evaluation Metrics} \label{evaluation_metrics}
We employ different evaluation metrics including 
the Mean Absolute Error (MAE) and  Percentage of 
Tick Accuracy (PTA) \cite{polyzou2016grade}. In the grading 
system, there are 11 letter grades (A+, A, A-, B+, B, B-, C+, C, C-, D, F) which correspond to (4, 4, 3.67, 3.33, 3, 2.67, 2.33, 2, 1.67, 1, 0). A tick is defined as the difference between two consecutive letter grades. The performance of a model is assessed based on how many ticks away the predicted grade is from the true grade. For example, a true grade of B vs. prediction of B is zero tick; true grade of  B vs prediction of 
B- is one tick and true grade of B vs a prediction of C+ is two ticks. Table \ref{tab:tick} shows an example. To assess the performance of the models by using PTA, we first convert the predicted numerical grades to the closest letter grades and then compute the percentages of each of the $x$ ticks.
\begin{table}[t!]
	\centering
	\caption{Tick error example} \label{tab:tick}
	\begin{tabular}{ccc}
	\hline
	\hline
	True Grade & Predicted Grade & Tick Error \\
	\hline
	B & B & = 0 \\
	\hline
	B & B-, B, B+ & $\leq$ 1 \\
	\hline
	B & C+, B-, B, B+, A- & $\leq$ 2 \\
	\hline
	\end{tabular}
\end{table}

\subsection{Comparative Methods}
We compare the proposed models with different approaches 
including matrix factorization and the traditional course-specific models.
% \vspace{-3mm}
\subsubsection{Bias Only (BO)}
The Bias Only method only takes into account a student's bias, course's bias and global bias \cite{hu2017enriching}. The predicted grade 
$\hat{g}_{s,c}$ by using this model is estimated as
\begin{equation}
\hat{g}_{s,c} = b_0 + b_s + b_c
\end{equation}
where $b_0$, $b_s$ and $b_c$ are the global bias, student bias and course bias, respectively.
% \vspace{-3mm}
\subsubsection{Matrix Factorization (MF)}
To use matrix factorization for a student's grade prediction, we assume that students and courses can be jointly represented in low-dimensional
latent space \cite{hu2017enriching}. 
% The students and courses in this model are represented by using latent vectors. 
The grade of student $s$ in a
future course $c$ can be predicted as
\begin{equation}
\hat{g}_{s,c} = b_0 + bs_s + bc_c + \bf{p}_s^T\bf{q}_c
\end{equation}
where $b_0$, $b_s$ and $b_c$ are the global bias, student bias and course bias, respectively; $\bf{p}_s$ and $\bf{q}_c$ are latent vectors corresponding to student $s$ and course $c$.

% \vspace{-3mm}
\subsubsection{Course-Specific Regression with Prior Courses}
The course-specific regression with prior courses (CSR\textsubscript{PC}) \cite{polyzou2016grade1} predicts the 
grade of a student $s$ in a course $c$ as a linear combination of the grades in prior courses. 
% The predicted grade $\hat{g}_{s,c}$ is 
% \begin{equation}
% \hat{g}_{s,c} = w_{c0} + \bf{x}_{s,c}^T\bf{w}_c^{pr}
% \end{equation}
% where $\bf{x}_{s,c} \in R^{m_c}$ is a feature vector encoding the grades of prior courses, $w_{c0}$ is the bias term and $\bf{w}_c^{pr} \in R^{m_c}$ is the weight vector to be learned and $m_c$ is the number of prior courses.
% \vspace{-3mm}
\subsubsection{Course-Specific Regression with Content Features}
The Course-Specific Regression with Content Features (CSR\textsubscript{CF}) model \cite{hu2017enriching} predicts a student's grade in a course using content features related to the student (e.g.,
academic level, previous GPAs, major, etc) and the course (e.g., course, discipline, credit hours, etc.). 
% For a full list of content 
% features, refer to \cite{hu2017enriching}.  The predicted
% grade is 
% \begin{equation}
% \hat{g}_{s,c} = w_{c0} + \bf{x}_{s,c}^T\bf{w}_c^f
% \end{equation} 
% where $\bf{x}_{s,c}$ is the content feature vector, $\bf{w}_c^f$ is the weight vector to be learned and $w_{c0}$ is the bias term.
% \vspace{-3mm}
\subsubsection{Course-Specific Hybrid Model}
The course-specific hybrid model (CSR\textsubscript{HY}) predicts a student's grade in a future course by
combining the content features and grades of prior courses \cite{hu2017enriching}.

\section{Results and Discussion}
%In this section we present the evaluation
%results on different evaluation metrics and the analysis of the results. We 
%also present the interpretations of the predictions made by the models.

\subsection{Comparative Performance}
\begin{table*} [h!]
	\centering
	\caption{Comparative Performance of different models using MAE. ($\downarrow$ is better)} \label{tab:a}
	\begin{adjustbox}{max width=\textwidth}
	\begin{tabular}{C{1.1cm} | C{1cm}C{1cm}C{1cm}C{1cm}C{1cm} | C{1cm}C{1cm}C{1cm}C{1cm}C{1cm} }
		\hline \hline
		\multirow{2}{3em}{Method} & \multicolumn{5}{c|}{Fall 2016} & \multicolumn{5}{c}{Spring 2017} \\\cline{2-11}
		& CS & ECE & BIOL & PSYC & CEIE & CS & ECE & BIOL & PSYC & CEIE \\
		\hline
		BO & 0.725 & 0.690 & 0.541 & 0.595 & 0.586 & 0.763 & 0.604 & 0.621 & 0.609 & 0.617 \\
		MF & 0.718 & 0.679 & 0.542 & 0.609 & 0.579 & 0.701 & 0.589 & 0.625 & 0.622 & 0.583 \\ 
		CS\textsubscript{MF} & 0.715 & 0.666 & 0.536 & 0.567 & 0.573 & 0.696 & 0.540 & 0.624 & 0.603 & 0.572 \\
		CSR\textsubscript{PC} & 0.680 & 0.673 & 0.537 & 0.493 & 0.601 & 0.666 & 0.506 & 0.566 & 0.567 & 0.517 \\
		CSR\textsubscript{CF} & 0.718 & 0.677 & 0.476 & 0.474 & 0.609 & 0.683 & 0.553 & 0.586 & 0.565 & 0.461 \\
		CSR\textsubscript{HY} & 0.669 & 0.663 & 0.505 & 0.485 & 0.583 & 0.663 & 0.502 & 0.560 & 0.558 & 0.465 \\
		\hline
		\textbf{MLP} & 0.590 & 0.450 & 0.429 & 0.353 & 0.395 & 0.606 & 0.368 & 0.517 & 0.491 & 0.419 \\
		\textbf{LSTM} & \textbf{0.588} & \textbf{0.367} & \textbf{0.412} & \textbf{0.316} & \textbf{0.324} & \bf{0.579} & \textbf{0.286} & \bf{0.500} & \textbf{0.392} & \textbf{0.253} \\
		\hline
	\end{tabular}
	\end{adjustbox}
\end{table*}

Table \ref{tab:a} shows the comparison of the proposed  MLP and LSTM models 
with various 
baselines using 
MAE for the Fall 2016 and Spring 2017 semesters. We observe 
that the deep learning models have the best performance 
on all datasets and LSTM outperforms MLP approach. Specifically, the 
LSTM model outperforms the best 
performing baseline by 12 to 45\%  across the different majors and the 
two semesters. 
%12\%, 44\%, 13\%, 33\% and 43\% on CS, ECE, BIOL, PSYC and CEIE, 
%respectively, in Fall 2016; and 12\%, 43\%, 10\%, 30\% and 45\% in Spring 2017. 
Compared to MLP, the LSTM model is able to achieve better performance. The reason is that LSTM can model the temporal dynamics associated with students' knowledge evolution, which can not be captured by MLP and other traditional methods. In addition, LSTM are able to handle long-term dependencies within the knowledge evolution. For example, an important course such as the prerequisite taken several semesters away can have a significant effect on the course to be predicted, which can be modeled by LSTM.

% \subsection{Performance on PTA}
\begin{table*} [th!]
	\centering
	\caption{Comparative Performance of Different Models using Tick Error ($\uparrow$ is better)}\label{tab:b}
	\begin{adjustbox}{max width=\textwidth}
	\begin{tabular}{C{1cm}|C{1cm}|C{1cm}C{1cm}C{1cm}C{1cm}C{1cm}|C{1cm}C{1cm}C{1cm}C{1cm}C{1cm}}
		\hline \hline
		 & \multirow{2}{3em}{} & \multicolumn{5}{c|}{Fall 2016} & \multicolumn{5}{c}{Spring 2017} \\\cline{2-12}
		 & Method & CS & ECE & BIOL & PSYC & CEIE & CS & ECE & BIOL & PSYC & CEIE \\
		\hline
		\multirow{8}{3em}{\textbf{$\text{PTA}_0$}} %Percentage of Grades predicted with no error
		& BO & 6.71 & 10.38 & 15.36 & 27.07 & 20.08 & 13.41 & 13.41 & 23.22 & 33.20 & 28.28 \\
		& MF & 7.11 & 13.11 & 15.90 & 28.34 & 21.26 & 18.49 &  17.07 & 23.36 & 33.21 & 27.87 \\
		& CS\textsubscript{MF} & 6.72 & 12.57 & 16.17 & 28.98 & 21.26 & 20.80 & 14.31 & 22.38 & 31.55 & 31.56 \\
		& CSR\textsubscript{PC} & 19.57 & 20.77 & 28.84 & 34.08 & 27.17 & 21.87 & 17.68 & 25.17 & 35.19 & 31.97 \\
		& CSR\textsubscript{CF} & 13.44 & 16.39 & 28.03 & 27.39 & 29.13 & 15.10 & 16.78 & 24.48 & 33.25 & 38.52 \\
		& CSR\textsubscript{HY} & 19.76 & 22.40 & 30.73 & 35.35 & 26.38 & 21.26 &  18.62 & 26.15 & 36.17 & 37.70 \\
		& \textbf{MLP} & 26.48 & 42.62 & 38.17 & 46.50 & 39.40 & 26.38 & 42.07 & 31.61 & 39.56 & 33.20 \\
		& \textbf{LSTM} & \bf{28.23} & \bf{50.27} & \bf{41.40} & \bf{50.85} & \bf{52.81} & \bf{28.88} & \bf{53.05} & \bf{35.92} & \bf{45.36} & \bf{54.92} \\
		\hline
		%Percentage of grades predicted with an error of at most one tick
		\multirow{8}{3em}{\textbf{$\text{PTA}_1$}} 
		& BO & 29.84 & 33.33 & 28.84 & 45.54 & 35.83 & 48.84 & 43.29 & 46.79 & 60.81 & 70.49 \\
		& MF & 29.84 & 31.15 & 29.65 & 43.95 & 34.25 & 48.69 & 41.46 &  47.69 & 60.80 & 68.03 \\
		& CS\textsubscript{MF} & 30.24 & 33.87 & 29.38 & 45.86 & 34.65 & 47.46 & 40.85 & 48.95 & 61.57 & 70.08 \\
		& CSR\textsubscript{PC} & 48.22 & 55.19 & 62.80 & 61.15 & 52.76 & 42.84 & 37.19 & 46.01 & 62.37 & 67.21 \\
		& CSR\textsubscript{CF} & 44.66 & 51.37 & 70.89 & 64.97 & 52.76 & 45.76 & 36.59 & 49.59 & 65.29 & 66.39 \\
		& CSR\textsubscript{HY} & 49.80 & 55.19 & 67.38 & 61.78 & 53.15 & 42.68  & 36.58 & 49.73 & 61.89 & 66.80 \\
		& \textbf{MLP} & 57.51 & 67.76 & 72.31 & 72.29 & 70.99 & 59.87 & 73.17 & 65.03 & 64.56 & 65.98 \\
		& \textbf{LSTM} & \bf{58.05} & \bf{78.69} & \bf{73.66} & \bf{77.97} & \bf{78.79} & \bf{60.71} & \bf{78.05} & \bf{67.04} & \bf{73.43} & \bf{84.84} \\
		\hline
		% Percentage of grades predicted with error of at most two ticks
		\multirow{8}{3em}{\textbf{$\text{PTA}_2$}}
		& BO & 60.67 & 58.47 & 62.80 & 69.75 & 58.66 & 66.56 & 62.80 & 60.22 & 78.54 & 82.79 \\
		& MF & 59.88 & 57.92 & 61.46 & 68.15 & 57.48 &  67.79 & 60.36 & 61.70 & 78.64 & 83.19 \\
		& CS\textsubscript{MF} & 61.26 & 59.02 & 61.19 & 71.97 & 57.48 & 67.80 & 60.36 & 61.79 & 80.97 & 83.61 \\
		& CSR\textsubscript{PC} & 74.31 & 73.22 & 81.40 & 79.62 & 69.69 & 65.02 & 57.31 & 63.98 & 77.18 & 84.02 \\
		& CSR\textsubscript{CF} & 73.52 & 75.96 & 87.87 & 83.44 & 66.14 & 65.95 & 56.71 & 62.54 & 78.91 & 80.10 \\
		& CSR\textsubscript{HY} & 75.10 & 74.32 & 82.75 & 78.66 & 69.29 & 63.64 & 57.31 & 63.84 & 77.69 & 81.97\\
		& \textbf{MLP} & 79.25 & 83.61 & \bf{87.90} & 86.31 & \bf{90.91} & 78.99 & 90.24 & 82.66 & 80.34 & 86.48 \\
		& \textbf{LSTM} & \bf{79.32} & \bf{88.52} & 87.63 & \bf{87.80} & 88.74 & \bf{79.97} & \bf{92.07} & \bf{83.66} & \bf{86.97} & \bf{92.62} \\
		\hline
	\end{tabular}
	\end{adjustbox}
\end{table*}

To gain better insights into 
the types of errors made by different methods, Table \ref{tab:b} presents
the experimental results evaluated by using tick errors as 
defined in Section \ref{evaluation_metrics}. The 
LSTM model achieves the best performance  
(with exceptions in BIOL and CEIE majors for Fall 2016). Similar to 
results evaluated by using MAE, MLP is inferior to LSTM but better 
than the other competing methods. We also observe that the
gap between the proposed 
methods and the baselines is smaller for PTA$_2$ that allows for errors up to two ticks to be counted as correct. The baseline
models with content features show better performance than 
methods that do not use content features. This suggests that 
content features are informative 
for performance prediction.

\subsection{Identifying At-Risk Students}

\begin{table*}%[t]
	\centering
	\caption{Predictive Power at Identifying At-Risk Students for Spring 2017 ($\uparrow$ is better)} \label{tab:d}
	\begin{adjustbox}{max width=\textwidth}
	\begin{threeparttable}
	\begin{tabular}{ C{1cm} | C{1cm}C{1cm} | C{1cm}C{1cm} | C{1cm}C{1cm} | C{1cm}C{1cm} | C{1cm}C{1cm} }
	\hline \hline
	Method & \multicolumn{2}{c|}{CS} & \multicolumn{2}{c|}{ECE} & \multicolumn{2}{c|}{BIOL} & \multicolumn{2}{c|}{PSYC} & \multicolumn{2}{c}{CEIE} \\
	\hline
	& Acc & F-1 & Acc & F-1 & Acc & F-1 & Acc & F-1 & Acc & F-1 \\
	\hline
	BO & 0.751 & 0.320 & 0.798 & 0.547 & 0.814 & 0.616 & 0.859 & 0.194 & 0.938 & 0.651 \\
	MF & 0.730 & 0.358 & 0.780 & 0.571 & 0.805 & 0.621 & 0.854 & 0.189 & 0.893 & 0.458 \\
	CS\textsubscript{MF} & 0.779 & 0.243 & 0.835 & 0.542 & 0.807 & 0.568 & 0.873 & 0.133 & 0.938 & 0.482 \\
	CSR\textsubscript{PC} & 0.773 & 0.369 & 0.823 & 0.452 & 0.806 & 0.560 & 0.868 & 0.181 & 0.889 & 0.228 \\
	CSR\textsubscript{CF} & 0.776 & 0.248 & 0.786 & 0.222 & 0.805 & 0.490 & 0.893 & 0.120 & 0.913 & 0.160 \\
	CSR\textsubscript{HY} & 0.779 & 0.375 & 0.829 & 0.517 & 0.808 & 0.559 & 0.864 & 0.200 & 0.922 & 0.296 \\
	\hline
	\bf{MLP} & 0.796 & 0.388 & 0.878 & 0.714 & 0.830 & \bf{0.636} & 0.859 & 0.236 & 0.938 & 0.516 \\
	\bf{LSTM} & \bf{0.819} & \bf{0.425} & \bf{0.920} & \bf{0.826} & \bf{0.833} & 0.624 & \bf{0.907} & \bf{0.274} & \bf{0.954} & \bf{0.666} \\
	\hline
	\end{tabular}
	\begin{tablenotes}
		\small
		\item In Spring 2017, the percentage of at-risk students for each major is CS (22.19\%), ECE (23.78\%), BIOL (26.01\%), PSYC (10.68\%), CEIE (8.61\%).
	\end{tablenotes}
	\end{threeparttable}
	\end{adjustbox}
\end{table*}

One of the important applications of student's performance 
prediction is to develop an early-warning system that 
can identify students at-risk of failing courses 
they plan to take in the next term (or future). 
%Therefore, the 
%predictive performance of an algorithm at catching at-risk students is 
%an important index. 
%To assess the capabilities of the proposed methods a
%at-risk student, we design the following experiments. 
We define at-risk students as those whose grades
are below 2.0 (C and lower). We convert all the grades 
above 2.0 as not-at-risk and below 2.0 as at-risk, and treat 
the prediction as a classification problem. The experimental procedures are 
similar to grade prediction except that the predicted grades over 2.0 
are treated as pass and below 2.0 as fail. We choose accuracy and F-1 score as evaluation
metrics, due to the fact that the number of at-risk and non-at-risk students is imbalanced.
% as indicated by the percentage of at-risk students at the footnote of Table \ref{tab:c} and \ref{tab:d}. 
The
experimental results are shown in Table
% Table \ref{tab:c} and 
\ref{tab:d} for Spring 2017. Higher value is better.
We observe that the proposed methods outperform 
all the baselines and in most cases, LSTM performs better than MLP.

\subsection{Uncertainty Evaluation}
\begin{figure}
	\centering
	\includegraphics[width=0.9\linewidth]{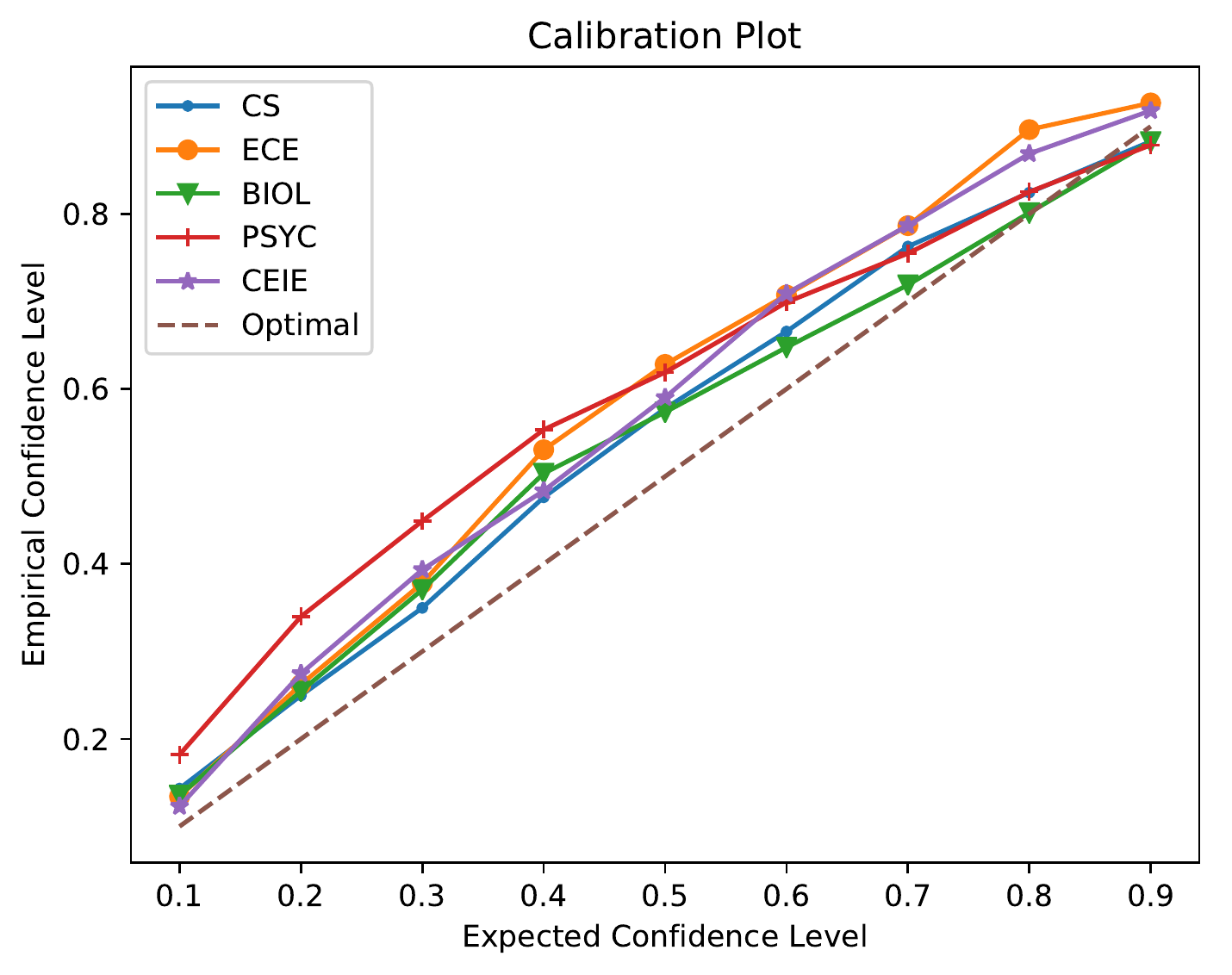}
	\caption{Empirical confidence level vs. expected confidence level} \label{fig:coverage}
\end{figure}

In this section, we evaluate the quality of our uncertainty estimation in several aspects. The first one is coverage, which can be evaluated by using calibration plot. The calibration plot can be explained by observing that if a prediction is made at $95\%$ confidence level, then 
the probability that the prediction falls in the prediction interval should be $0.95$.  Figure \ref{fig:coverage} shows the 
calibration plot evaluated on our datasets. The x-axis of the plot is the expected confidence level and the y-axis is the empirical confidence level. From the figure, we can see that the calibration curves across 
the five majors are close to the optimal calibration curve.

\begin{figure}
	\centering
	\includegraphics[width=0.9\linewidth]{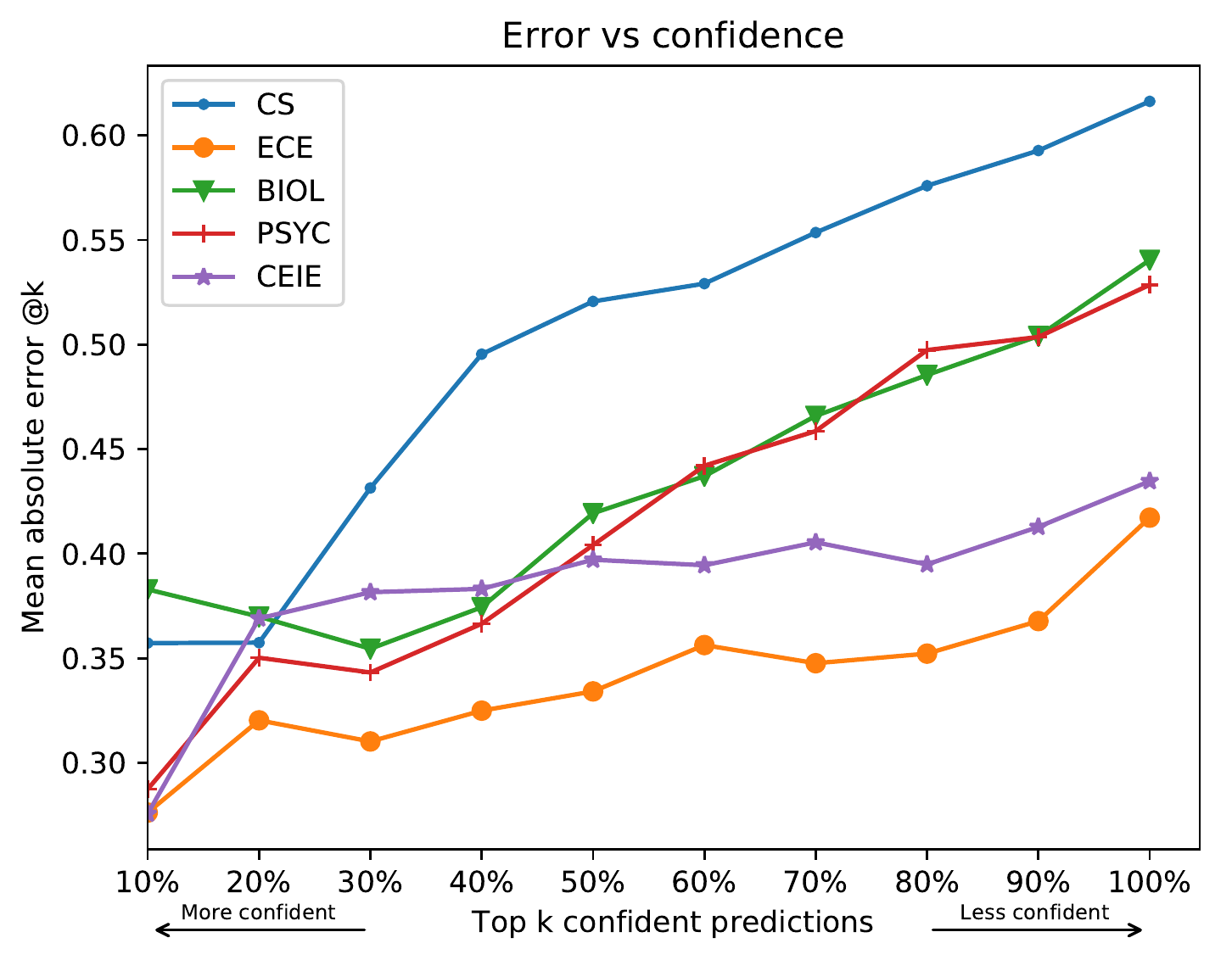}
	\caption{MAE as a function of confidence.} \label{fig:error_confidence}
\end{figure}

The second way to evaluate uncertainty estimation is that a model should make less errors on predictions that it is confident about. Therefore, we evaluate the model as a function of confidence score and we propose error@k:
\begin{equation*}
	E@k = \text{error}(\text{k most confident predictions}),
\end{equation*}
where error can be mean absolute error or tick error, the predictions are ranked in terms of predictive variance, lower predictive variance is more confident. Figure \ref{fig:error_confidence} shows mean absolute error with respect to top-k confident predictions. The figure shows that as the predictions become less confident, the prediction errors become higher.

\begin{figure}[h]
	\centering
	\begin{subfigure}[b]{.9\columnwidth}
		\centering
		\includegraphics[width=\linewidth]{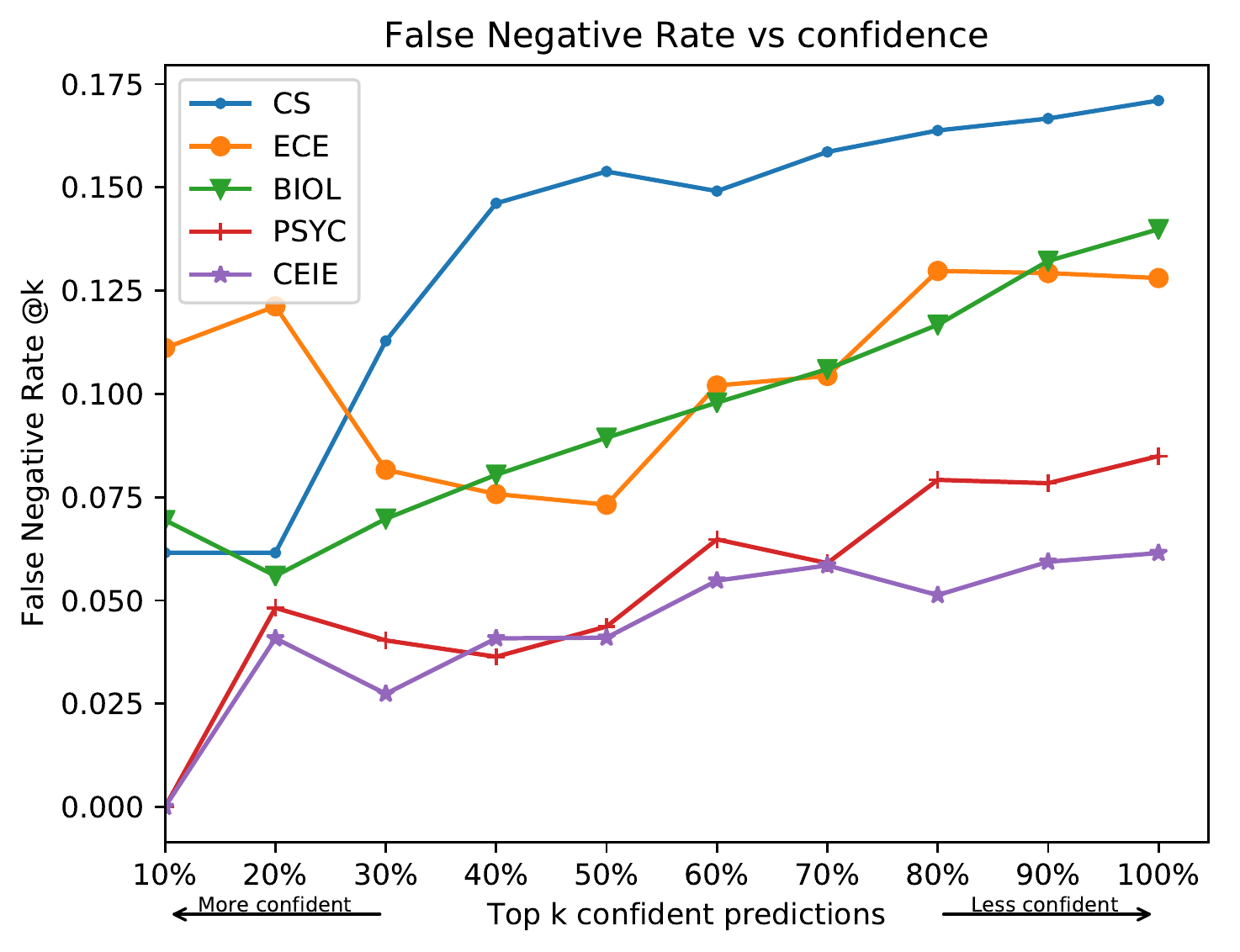}
		\caption{FNR vs confidence} \label{fig:fnr}
	\end{subfigure}
	\vfill
	\begin{subfigure}[b]{.9\columnwidth}
		\centering
		\includegraphics[width=\linewidth]{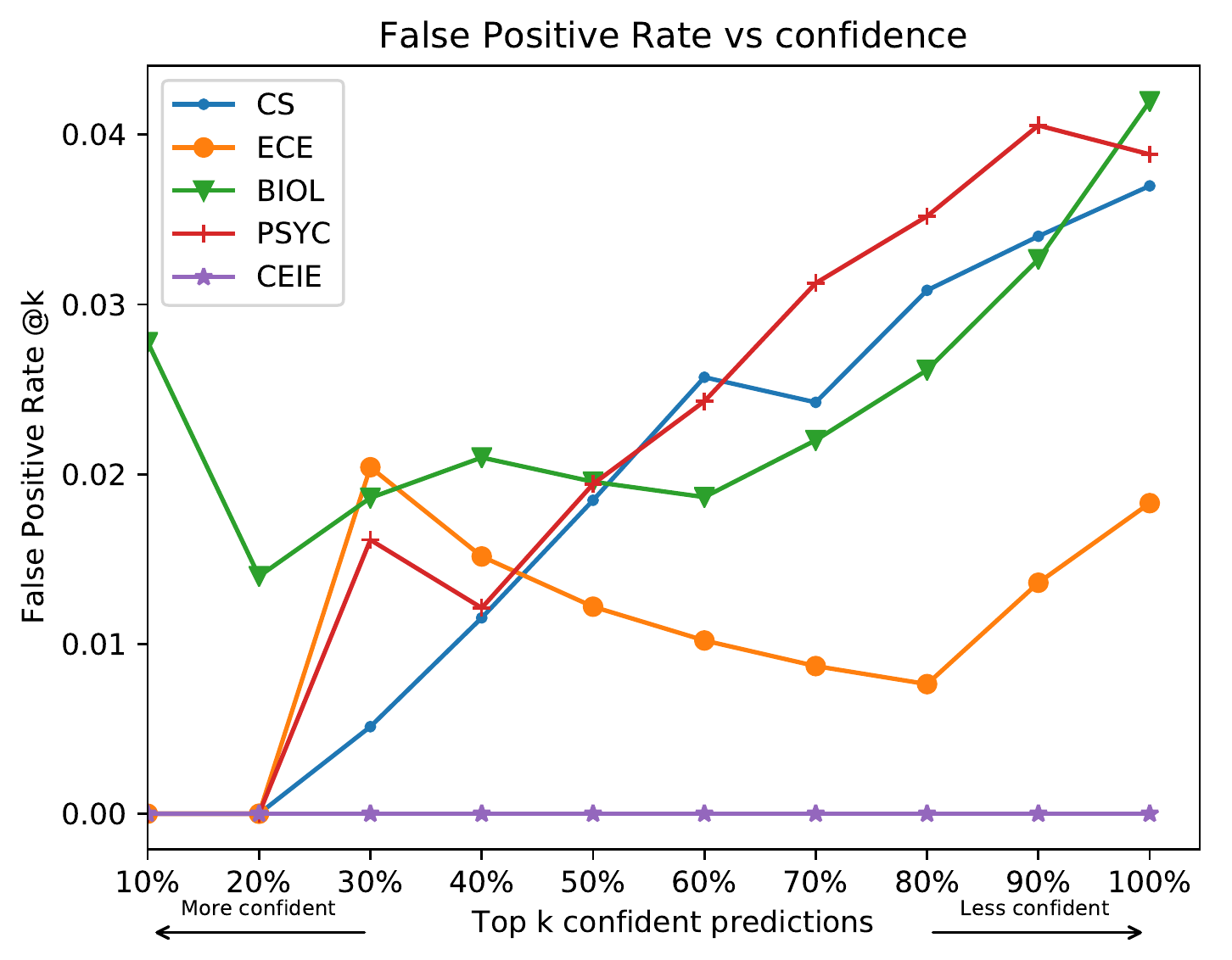}
		\caption{FPR vs confidence} \label{fig:fpr}
	\end{subfigure}
	\caption{FNR and FPR as a function of prediction confidence} \label{fig:fnr_fpr}
\end{figure}

Grade prediction is fundamental for  early-warning student facing 
system. The application case of educational early-warning system is that when an instructor/advisor 
is informed that a student will fail a course, the 
instructor will reach out to the student and provide the student personalized advising and help. In this process, we want to make sure students who are not failing are  
not predicted as failing students (false positives) so as to reduce wasted educational 
resources; and the failing students are  
not predicted as passing 
students (false negatives) so that they can get much needed timely help. Prediction confidence 
is useful for reducing these kind of errors by only taking action on confident predictions. We also evaluate
uncertainty estimation on the application of identification of at-risk students. Figure \ref{fig:fnr_fpr} shows 
false negative rate (FNR) and false positive rate (FPR) as a function of prediction confidence.

We observe in most cases as prediction confidence decreases, there are more errors. To make less errors, we can choose 
to take actions only on confident predictions. However, more confident predictions have less coverage. In practice, we propose to 
set an
appropriate confidence threshold to make a tradeoff between coverage and accuracy. All the uncertainties are estimated by using MLP and we levae uncertainty estimation using LSTM to future work.

\subsection{Case Studies: Influential Courses}
\begin{figure*}
	\centering
	\includegraphics[width=0.8\linewidth]{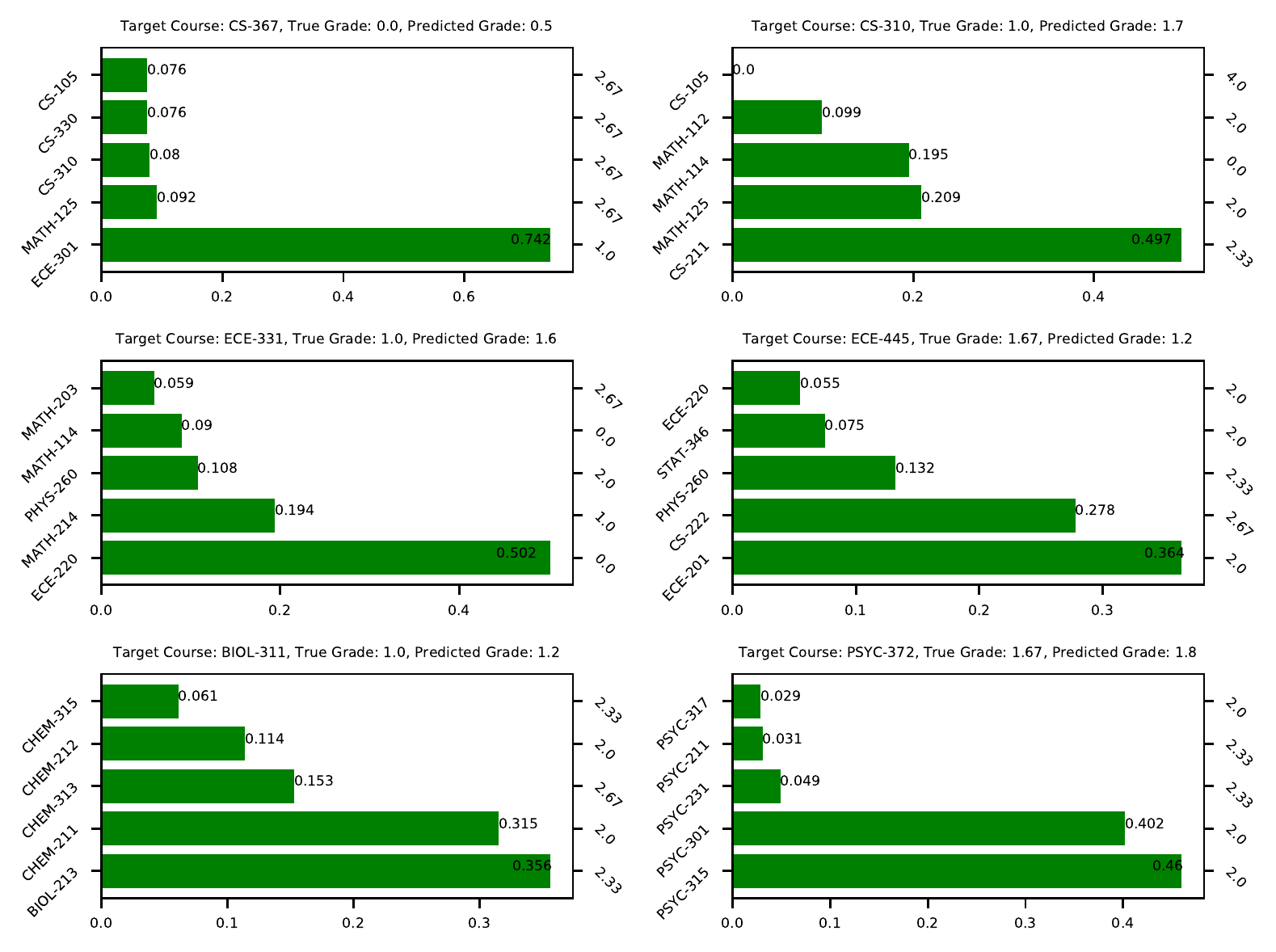}
	\caption{Influence of prior courses. For every subfigure, x-axis is influence value, left y-axis is top five most influential prior courses, right y-axis is student's grade in corresponding prior course. In the titles, target course means the course for which we are predicting grade, predicted grade is the predicted grade for the student, and true grade is the student's real grade in the target course.} \label{fig:inter}
\end{figure*}

To incorporate the developed next-term grade approaches within 
personalized advising system, we seek to not only report the 
predicted grades for the student but identify the list of 
prior courses that were most influential for determining future 
success in a given target course.

Figure \ref{fig:inter} shows examples of use case scenarios 
for students in different disciplines.  We choose six at-risk 
students from CS, ECE, BIOL and PSYCH 
majors. If a student has grade lower than 2.0, 
he/she is identified as at-risk student.
We compute the influence of the prior 
courses on the prediction as described in Section \ref{explain} 
and only prior courses contributing to the increase in the 
prediction are reported. The influence is computed by using the 
LSTM model. The influence index is sorted and normalized and 
only the top five prior courses are shown in the table. 
% For the details of the courses, please refer to GMU catalog \cite{gmu:catalog}.
%
% From the tables, we can see that the students are at-risk and the model predicts them to be at-risk of failing the corresponding courses. 
% The influence index can help explain which prior courses lead to the prediction of failure. 
From the top left subfigure, we can see that the student's true 
grade in class CS-367 (this course is about computer systems and programming) is 0.0, the predicted grade is 0.5 and the most 
influential prior course is ECE-301 (class about digital electronics) which is a prerequisite. 
%; ECE-301 is the 
%prerequisite for this course. 
The influence of a prior course is computed by increasing 
the grade of that course to full grade (i.e. 4.0). This does not 
suggest that a course with a lower grade has higher influence than a course with higher grade. 
For example, in top right subfigure, the lowest 
grade of the student is from MATH-114 (about calculus), but the most influential 
is course  CS-211 (about object-oriented programming) (prerequisite of the target course). 
The left column of second row shows the third example. The target course is ECE-331 (about digital system design). For this 
student, the 
actual grade in the target course is 1.0 and predicted grade is 1.6. The most influential course is ECE-220 (about signals and systems), as 
this student performed very poorly in this course (grade was 0.0).
From the results, we can see that the proposed approach is able to identify 
influential prior courses that explain the prediction results.

\begin{table}[h!]
	\centering
	\begin{adjustbox}{max width=\linewidth}
	\begin{tabular}{|c|c|c|c|c|c|}
		\hline
		Target Course & \multicolumn{5}{c|}{Top 5 Influential Prior Courses} \\
		\hline
		CS-367 & ECE-301 & CS-310 & MATH-125 & \textbf{CS-262} & MATH-213 \\
		\hline
		ECE-433 & \textbf{ECE-333} & MATH-203 & STAT-346 & PHYS-160 & ENGR-107 \\
		\hline
		PSYC-372 & PSYC-325 & PSYC-301 & PSYC-300 & PSYC-211 & \textbf{PSYC-100} \\
		\hline
		BIOL-311 & BIOL-213 & CHEM-211 & CHEM-212 & CHEM-313 & \textbf{BIOL-214} \\
		\hline
		CEIE-331 & PHYS-261 & \textbf{CEIE-210} & STAT-344 & PHYS-161 & MATH-113 \\
		\hline
	\end{tabular}
	\end{adjustbox}
	\caption{Top 5 influential prior courses for a target course. Bolded course is the pre-requisite.} \label{tab:collective_inf}
\end{table}

Table \ref{tab:collective_inf} shows the influence of prior courses on a target course for groups of students. We choose 
five representative courses from each major. From the table, we observe that for all the courses their 
prerequisites are one of the top five most influential courses. Although, most of the time, the most 
influential prior course for a target course is not necessarily the prerequisite, the most influential 
course is very relevant to the target course. For example, course CS-367 about 
low-level computer system such as machine-level programming; the most influential 
course ECE-301 is digital electronics, which is about designing logic circuits and relevant to low-level 
computer system. 
Providing a list of influential courses for a target course can help stakeholders 
improve the curriculum and program structure.
%
%
% In the appendix, we provide course names about the courses mentioned in this section.

\section{Conclusions}
In this work, we proposed two course-specific Bayesian deep learning models for next-term
grade prediction. The first  is a multi-layer perceptron 
which treats the feature vector as static and 
ignores the temporal dynamics of a 
student's knowledge evolution. To overcome this issue, we also developed a Long Short Term Memory model 
that takes into account the sequential aspect of student's knowledge accumulating process by taking courses across semester/terms.

We highlight the strengths of our proposed approach by incorporating the predictions within three application 
scenarios: (i) identify at-risk students and 
(ii) provide explainable results so as to identify a list of influential courses associated with a target course.
(iii) provide prediction uncertainty for building a reliable educational early warning system.

We conducted comprehensive experiments to evaluate the proposed models. The experiments demonstrate 
that the proposed models exhibit better performance at predicting students' grades than state-of-the-art baselines.  The experiments also show 
that the proposed models have better capability at identifying at-risk students.

% \begin{appendices}
% \section{Course Names}

% \begin{table} [h!]
%     \centering
%     \begin{adjustbox}{max width=\linewidth}
%     \begin{tabular}{|C{1.1cm}|C{1cm}|C{1cm}|C{1cm}|}
%     \hline
%     Course Number & Course Name & Course Number & Course Name \\
%     \hline
%     BIOL-213 & Cell Structure and Function & BIOL-311 & General Genetics \\
%     \hline 
%     \end{tabular}
%     \end{adjustbox}
% \end{table}

% \end{appendices}

\begin{acks}
This work was supported by the National Science Foundation grant \#1447489. The computational resources was provided by ARGO, a research computing cluster provided by the Office of Research Computing at George Mason University, VA. (URL:http://orc.gmu.edu)
\end{acks}

\bibliographystyle{ACM-Reference-Format}
\bibliography{refs}

\end{document}